\documentclass[11pt,a4paper]{article}
\usepackage[hyperref]{emnlp-ijcnlp-2019}
\usepackage{times}
\usepackage{latexsym}

\usepackage{url}

\usepackage{amsmath}

\usepackage{graphicx}

\usepackage{url}
\usepackage{booktabs}

\usepackage{comment}

\usepackage[caption=false]{subfig}

\newcommand{\vinod}[1]{{}}

\newcommand{\fromm}[1]{{}}

\def\perturbdev{\textit{ScoreDev}}
\def\perturbrange{\textit{ScoreRange}}
\def\perturbjaccard{\textit{LabelDist}}
\def\perturbsens{\textit{ScoreSens}}

\aclfinalcopy 

\title{Perturbation Sensitivity Analysis to Detect Unintended Model Biases}

\author{Vinodkumar Prabhakaran \\
  Google Brain \\
  San Francisco, CA, USA \\
  {\tt \small vinodkpg@google.com} \\\And
  Ben Hutchinson \\
  Google Brain \\
  San Francisco, CA, USA \\
  {\tt \small benhutch@google.com} \\\And    Margaret Mitchell \\
  Google Brain \\
  Seattle, WA, USA \\
  {\tt \small mmitchellai@google.com} \\}

\date{}
\begin{document}
\maketitle
\begin{abstract}

Data-driven statistical Natural Language Processing (NLP) techniques leverage large amounts of language data to build models that can understand language. However, most language data reflect the public discourse at the time the data was produced, and hence NLP models are susceptible to learning incidental associations around named referents at a particular point in time, in addition to general linguistic meaning. An NLP system designed to model notions such as sentiment and toxicity should ideally produce scores that are independent of the identity of such entities mentioned in text and their social associations. For example, in a general purpose sentiment analysis system, a phrase such as \textit{I hate Katy Perry} should be interpreted as having the same sentiment as \textit{I hate Taylor Swift}. Based on this idea, we propose a generic evaluation framework, {\em Perturbation Sensitivity Analysis}, which detects unintended model biases related to named entities, and requires no new annotations or corpora. We demonstrate the utility of this analysis by employing it on two different NLP models --- a sentiment model and a toxicity model --- applied on online comments in English language from four different genres.

\end{abstract}

\section{Introduction}
\label{sec_intro}

Recent research has shown ample evidence that data-driven NLP models may inadvertently capture, reflect and sometimes amplify various social biases present in the language data they are trained on \cite{bolukbasi2016man, blodgett2017racial}.
Such biases can often result in unintended and disparate harms to the users who engage with NLP-aided systems.
For instance, when NLP algorithms are used to moderate online communication, e.g.\ by detecting harassment,
although the net social benefits may be positive, the harms caused by incorrect 
classifications may be 
unevenly
distributed, leading to disparate impact
\cite{feldman2015certifying}. 
Some writers may find their 
contributions being disproportionately censored, while some readers may not be adequately protected from harassment \cite{dixon2018measuring}.

\begin{table}[t]
\centering
\small
\begin{tabular}{@{}lcc@{}}
\toprule

Sentence & Toxicity & Sentiment \\
\midrule
I hate Justin Timberlake. & 0.90 & -0.30 \\
I hate Katy Perry. & 0.80 & -0.10 \\
I hate Taylor Swift. & 0.74 & -0.40 \\
I hate Rihanna. & 0.69 & -0.60 \\

\bottomrule
\end{tabular}

\caption{\small Sensitivity of NLP models to named entities in text. Toxicity score range: 0 to 1; Sentiment score range: -1 to +1.}
\label{table_problem}
\end{table}

Research into fairness in machine learning distinguishes
two broad categories of unfair discrimination. First, unfairness for {\em individuals}
exists when 
similar individuals are treated dissimilarly \cite{dwork2012fairness}.
Second, a range of 
criteria define unfairness for {\em groups}, each in terms of statistical 
dependence between group membership, model score, and class label (see, e.g., \cite{chouldechova2018frontiers,MitchellEtAl2018}).
In both cases, what is ``fair'' or ``unfair'' is highly context-dependent, and 
judgments about fairness require consideration of the broader sociotechnical frame \cite{selbst2019}.

This framework also poses some practical challenges:  individual fairness requires knowing intricate details about an individual, while group fairness  requires knowing how an individual can be categorized into legally and socially sensitive roles.  The first runs into the ethical concerns of surveillance; the second runs into the ethical concerns of discrimination. Furthermore, texts are often not annotated with the social groups of their readers/writers (and for privacy reasons we may not wish to infer them), or whether two individuals are ``similar'' or not.
Hence, fairness research in NLP has mostly focused on mentions of social identities \cite{dixon2018measuring, borkan2019nuanced, garg2019counterfactual}, or on how social stereotypes impact semantic interpretation \cite{webster2018mind}, and often rely heavily on annotated corpora.

In this paper, we propose a general-purpose evaluation framework that detects unintended biases in NLP models around named entities mentioned in text.
Our method does not rely on any annotated corpora, and we focus solely on application-independent 
{\em sensitivity} 
of models, which does not clearly fall under individual- or group- based fairness criteria. 
Our core idea is based on the assumption that an NLP system designed to be widely applicable should ideally produce scores that are independent of the identities of named entities mentioned in the text. For instance, the sentences {\em I hate Justin Timberlake} and {\em I hate Rihanna} 
both express the same semantics using identical constructions; however, 
the toxicity model used in our experiments gives a significantly higher score to the former ($0.90$) than the latter ($0.69$) (see Table~\ref{table_problem} for more examples).

\vinod{spell out what we do clearly; 
We perform our study in the English language, but design our approach so that it might be easily extended to other languages. mention pronoun.}

\vinod{move to last?}
Mentions of such real-world entities are pervasive in data. Just as word co-occurrence metrics capture `meaning representations' of words in the language,\footnote{Often through word embeddings fed to or learned by the first layer of neural network based models} co-occurrence patterns between entity mentions and other parts of the phrases they occur in influence their learned meaning. For example, if a person's name is often mentioned in negative linguistic contexts, a trained model might inadvertently associate negativity to that name, resulting in biased predictions on sentences with that name. 
If unchecked, this leads to undesirable biases in the model, violating tenets of both individual and group fairness as they are applied in context.

The primary contributions of this paper are:
    (i) a simple and effective general-purpose model evaluation metric, which we call {\em perturbation sensitivity analysis},
    for measuring unintended bias; 
    (ii) a large-scale systematic analysis of model sensitivity to name perturbations, on two tasks -- sentiment and toxicity -- across four different genres of English text; 
    (iii) a demonstration of how name perturbation can reveal undesired biases in the learned model towards names of popular personalities; 
    (iv) showing the downstream impact of name sensitivity, controlling for prediction thresholds. 

\section{Perturbation Sensitivity Analysis}
\label{sec_method}

We introduce \textit{Perturbation Sensitivity Analysis (PSA)}, a general evaluation technique to detect unintended biases in NLP models towards real-world entities.
Central to our approach is the notion of {\em perturbation}, where a reference to a real-world entity is replaced by a reference to another real-world entity of the same type (e.g., a person name replaced with another person name).
PSA measures the extend to which a model prediction is sensitive to such perturbations, and is calculated w.r.t. a set of (unannotated) sentences $X$ from the target domain and a set of names $N$ (of the same entity type $t$) that the perturbations are performed over.

For simplicity, in this paper, we discuss text classification models that take in a piece of text and return a score for a target class. Similarly, we focus on perturbing over person names. However, our method is readily extendable to other kinds of models as well as to other entity types.

Our approach begins by first retrieving the set of 
sentences $X$ 
such that each sentence contains at least one referring expression that refers to an entity of the type we are doing perturbation on (person, in our case). 
This referring expression could be a pronoun or a proper name.
We select one such referring expression as the \textit{anchor} for each sentence in $X$.
We then ``perturb'' each sentence by replacing the anchor with named entities $n \in N$.
We then measure the sensitivity of the model with respect to such perturbation by running it on the resulting set of $|X|*|N|$ perturbed sentences.

Formally, let $x_n$ denote the perturbed sentence obtained by replacing the anchor word in $x \in X$ with $n$, and 
$f(x_n)$
denote the score assigned to a target class by model $f$ on the perturbed sentence $x_n$. Formally, we define
three metrics for the {\em perturbation sensitivity of model scores}:

\paragraph{Perturbation Score Sensitivity} (\perturbsens) of a model $f$ with respect
 to a corpus $X$ and a name $n$
 is the average difference between $f(x_n)$ and $f(x)$
 calculated over $X$, i.e. $\underset{x \in X}{E}[f(x_n) - f(x)]$. 
 \vspace{-1mm}
\paragraph{Perturbation Score Deviation} (\perturbdev) of a model $f$ with  respect to a corpus $X$ and a set of names $N$ is
the standard deviation of scores due to perturbation, averaged across sentences, i.e.,
$\underset{x \in X}{E}[\underset{n \in N}{StdDev}(f(x_n)]$.
\vspace{-1mm}
\paragraph{Perturbation Score Range}  (\perturbrange)
of a model $f$ with  respect
to a corpus $X$ and a set of names $N$ is the
$Range$ ($max-min$) of scores, averaged  across sentences, i.e.,
$\underset{x \in X}{E}[\underset{n \in N}{Range}(f(x_n)]$.

\vspace{2mm}
\noindent Whether a score difference caused by name perturbation results in a different label depends also on the threshold. 
Given a threshold, $0 \leq c \leq 1$, binary labels $y(x)$ can be obtained from the classifier $f$ as $\textbf{I}[f(x)  \geq c] \in \{0, 1\}$, where $\textbf{I}[\cdot]$ is the indicator function.
Using this, we define a metric for the {\em perturbation sensitivity of model labels}:
\paragraph{Perturbation Label Distance} (\perturbjaccard)
of a binary classifier $y$ with  respect
to a corpus $X$ and a set of names $N$ is the
Jaccard Distance between a) the set of sentences $\{x\}$ for which $y(x)=1$,
and b) the sentences $\{x\}$ for which $y(x_n)=1$, averaged across names $n \in N$; i.e.,
$\underset{n \in N}{E}[Jaccard(\{x|y(x)=1\}, \{x|y(x_n)=1\})]$,\\
where $Jaccard(A, B) = 1-|A \cap B | / | A \cup B|$.

\subsection{Assumptions}

The underlying assumption of PSA is that the model should ideally be \textit{not} sensitive to name perturbation.
However, this assumption may not always hold true. Proper names \textit{do} convey meaning akin to the linguistic meanings expressed in more general phrases, and thus perturbing names may sometimes remove critical semantic content that an NLP system should be modelling. For example, \textit{he is like Hitler} vs. \textit{he is like Gandhi} should have very different sentiment scores, since the sentences evoke the pragmatics associated with those referents.
Whether the PSA assumption holds in individual sentences will depend on the sentential context; however, the corpus-level trends that we measure in the model scores/labels are still indicative of systemic biases in the model.
This points to the importance of paying care to how the corpus $X$ is constructed, and making sure that it captures a diverse set of sentential contexts.

\subsection{Analysis Framework}
\label{label_framework}

The PSA framework described above is applicable to any text classification models, on any target corpus, to detect bias with respect to any type of named entities (i.e., perturbable among each other). 
In this paper, we focus on two text classification models, applied to 4 different corpora, to detect biases associated with person names. 

\vspace{-2mm}

\paragraph{Models:}
We use two text classification models: a) a toxicity model that returns a score between [0,1], 
and b) a sentiment model that returns a score between [-1,+1].
Both models were trained using state-of-the-art neural network algorithms, and perform competitively on benchmark tests.\footnote{To obtain information about the models, for instance to perform replication experiments, please contact the authors.}

\vspace{-2mm}

\paragraph{Corpora:}
We use four socially and topically diverse corpora of online comments released by \newcite{voigt2018gender}: Facebook comments on politicians' posts (FB-Pol.) and on public figures' posts (FB-Pub.), Reddit comments, and comments in Fitocracy forums. For each corpus, we select 1000 comments at random that satisfy two criteria: at most 50 words in length, and contain at least one English 3rd person singular
pronouns (i.e., anchors). 
We use these extracted comments to build templates, where the pronouns can be replaced with a person name to form a grammatically coherent perturbed sentence. 
We use pronouns as the anchors for multiple reasons.
Pronouns are often {\it closed-class} words across languages,\footnote{While the assumption that pronouns are a closed-class is useful for many languages, Japanese and Malay are example languages where this assumption does not hold.} making it a useful reference cross-linguistically.
Using a list of names to query for anchors is an option; but it has the risk of resulting in a set of sentences biased towards the cultural/demographic associations of those names, a confound that the use of pronouns as anchors will avoid. 
We balance the representation of female and male pronouns in our extracted sentences so as to minimize the effect of skew towards one gender in particular within the test set. However future work should examine how to best account for non-binary genders in this step.

\begin{figure*}
    \centering
    \vspace{-1em}
    \includegraphics[width=.68\linewidth]{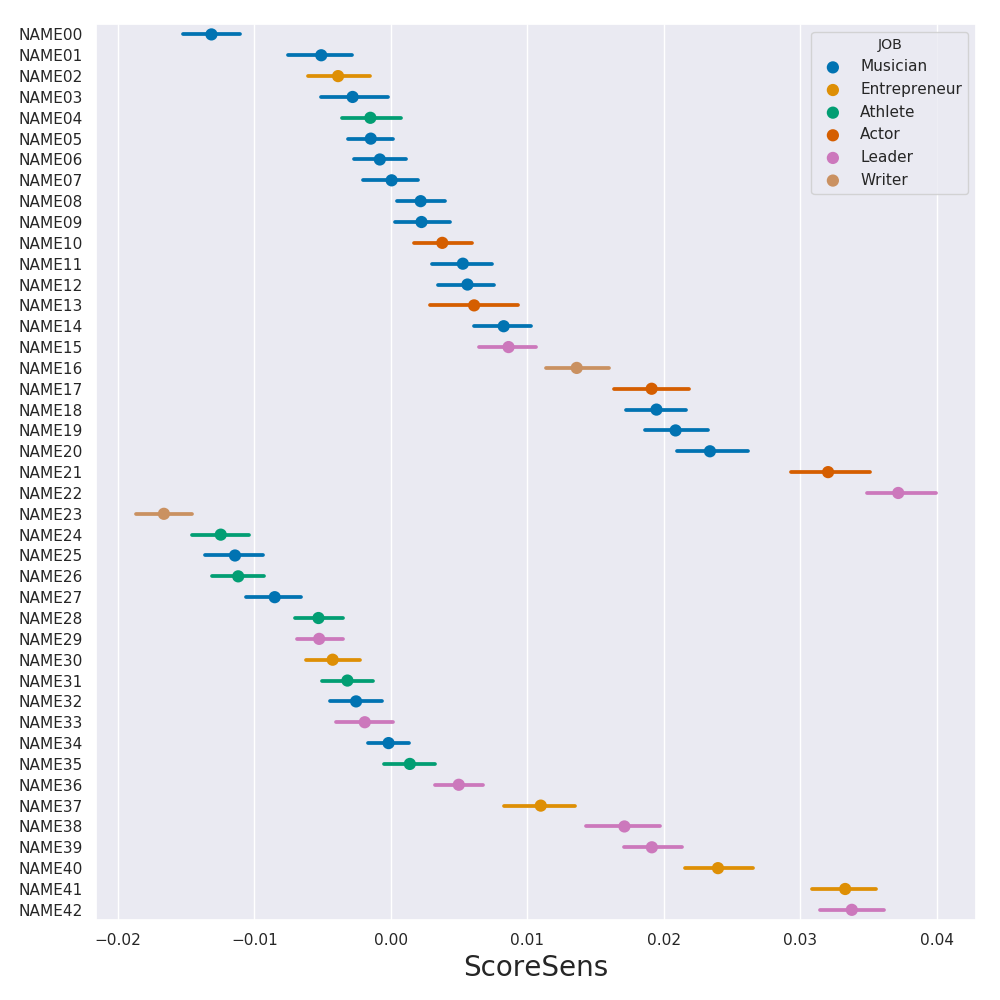}
    \caption{\small Name Perturbation Sensitivity (\perturbsens{}) for the toxicity model on the Reddit subcorpus, across names of controversial personalities.
    Female names are at the top; male names at the bottom; colors distinguish their career type.
    Names have been obfuscated due to their sensitive nature.
    }
    \label{fig:name_bias}
\end{figure*}

\vspace{-2mm}

\paragraph{Names:} 
We choose a list of controversial personalities, compiled based on Wikipedia page edit frequency.\footnote{\url{https://anon.to/x9PMYo}} Because of their controversial nature, these names are more likely to have social biases associated with them, which is helpful to demonstrate the utility of our analysis.

\vspace{2mm}

\section{Results}

Table~\ref{tab_numbers} shows the results of perturbation sensitivity analysis on different corpora. Both models exhibit significant sensitivity towards name perturbation across all 4 corpora. On average, sentences subjected to name perturbation resulted in a wide range of scores; i.e., \perturbrange{} over 0.10 for toxicity, and 0.36-0.42 for sentiment. 
Similarly, \perturbdev{} values for the sentiment model is also higher (over 0.07 across board) compared to that of the toxicity model (around 0.02), suggesting that the sentiment model is much more sensitive to the named entities present in text than the toxicity model.
We also observe that perturbation sensitivity is a function of the target corpus; comments on public figures had a much larger \perturbdev{} and \perturbrange{} for both tasks.

\begin{table}[t]
\centering
\small
\setlength{\tabcolsep}{3.5pt}
\begin{tabular}{@{}lcccc@{}}
\toprule
&\multicolumn{2}{c}{Toxicity}&\multicolumn{2}{c}{Sentiment} \\
Corpus & \perturbdev{} & \perturbrange & \perturbdev & \perturbrange  \\
\midrule
FB-Pol. & 0.022 & 0.107 & 0.070 & 0.360\\
FB-Pub. & 0.025 & 0.118 & 0.083 & 0.420 \\
Reddit & 0.022 & 0.107 & 0.072 & 0.376 \\
Fitocracy & 0.022 & 0.103 & 0.071 & 0.364 \\
\bottomrule
\end{tabular}
\caption{\small 
\perturbdev{} is the per-sentence standard deviation of scores upon name perturbation, averaged across all sentences.
\perturbrange{} is the per-sentence range of scores (i.e., max - min) upon name perturbation, averaged across all sentences.}
\label{tab_numbers}
\end{table}

\subsection{Bias Towards Specific Names}

We now analyze the \perturbsens{} for specific names. Figure~\ref{fig:name_bias} shows the \perturbsens{} for each name in our list, for the Toxicity-Reddit combination. 
Names are obfuscated in the figure due to their sensitive nature, but their career type is distinguished.
Replacing a pronoun with some names increases the toxicity scores by over $0.03$ on average, while other names decrease the scores by almost $0.02$ on average. 
It is also notable that leaders (politicians) and actors in our list have higher toxicity associations than musicians and athletes.
Similar effects also occur in the sentiment analysis model.

\subsection{Threshold Analysis}

Whether a score difference caused by perturbation results in a different label or not depends also on the threshold. 
It is possible that a model would be more stable on sentences with highly toxic language, but the effect of perturbation is more prevalent in sentences that have fewer signals of toxicity. We verified this to be the case in our analysis: the average (over all names) value of the  perturbation score sensitivity,
i.e. $|f(x_n) - f(x)|$, has a significant moderate negative correlation (-0.48) with the original score of that sentence, $f(x)$. This finding is of importance to counter-factual token fairness approaches such as \cite{garg2019counterfactual}.

\begin{figure}[t]
    \centering
    \includegraphics[trim={0 .62cm 0 0},clip,width=.92\linewidth]{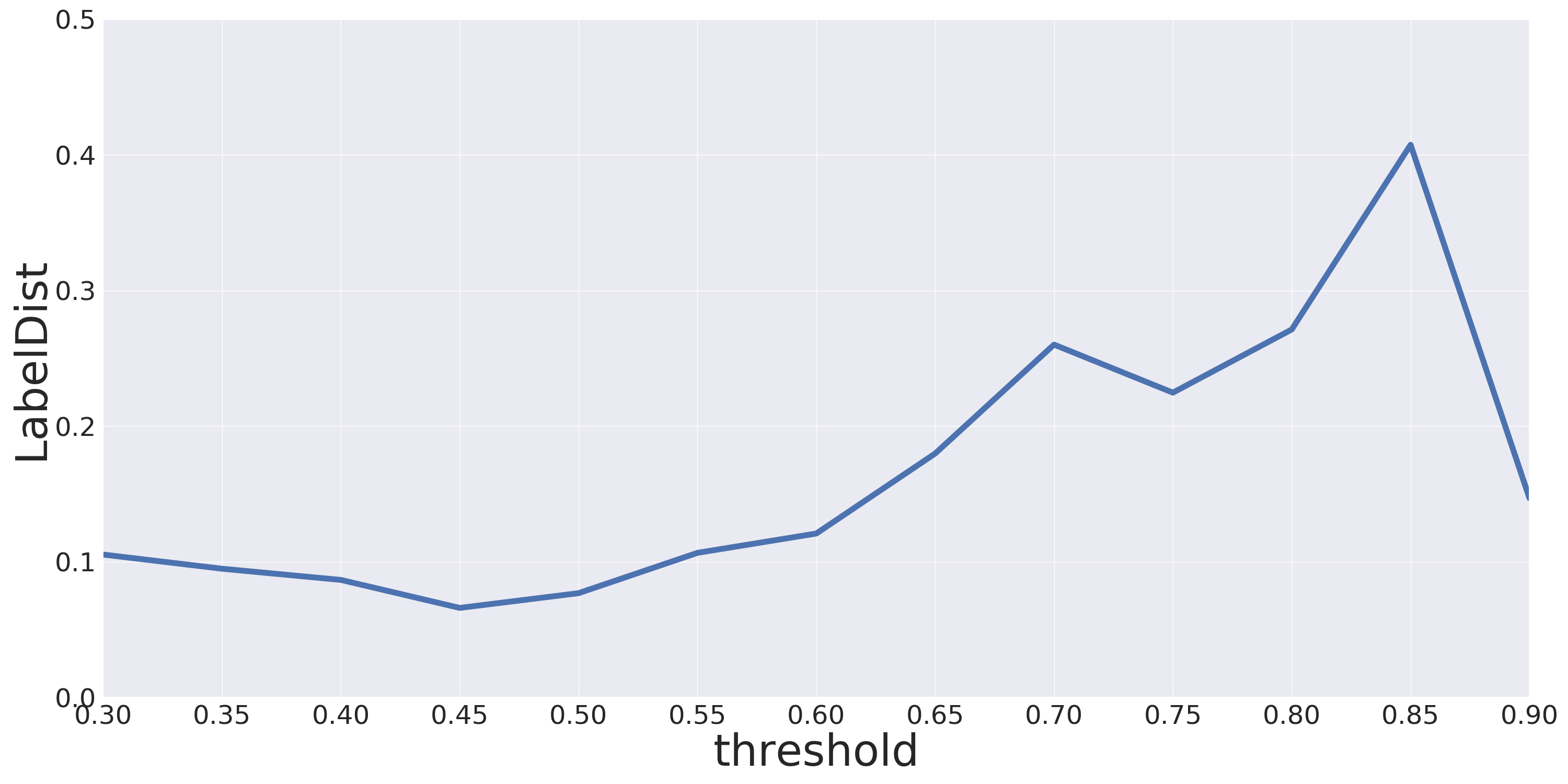}
    \caption{\small Even for high model thresholds, we see significant name perturbation sensitivity in classifications/labels.
    \perturbjaccard{} measures
    the \# of flips between \textit{toxic} and \textit{non-toxic}.
    }
    \label{fig:impact}
\end{figure}

To further understand the impact of perturbation sensitivity, we calculate \perturbjaccard{}, which takes into account the number of sentences that switch either from toxic to non-toxic 
or vice versa, when a pronoun is changed to a name. Figure~\ref{fig:impact} shows \perturbjaccard{} values across different thresholds. As can be seen from the Figure, the name perturbation results in a \perturbjaccard{} of 0.10 -- 0.40 across thresholds. 
This roughly suggests that around 10-40\% of sentences (with third person singular pronouns) labeled as toxic at any given threshold could flip the label as a result of name perturbation. It is also interesting to note that despite the negative correlation between $|f(x_n) - f(x)|$ and $f(x)$, the \perturbjaccard{} has high values at high thresholds.

\section{Related Work}
\label{sec_related}

Fairness research in NLP has seen tremendous growth in the past few years (e.g., \cite{bolukbasi2016man,caliskan2017semantics,webster2018mind,diaz2018addressing,dixon2018measuring,de2019bias,gonen-goldberg-2019-lipstick-pig,manzini-etal-2019-black}) over a range of NLP tasks such as co-reference resolution and machine translation, as well as the tasks we studied --- sentiment analysis and toxicity prediction. 
Some of this work study bias by swapping names in sentence templates \cite{caliskan2017semantics,kiritchenko-mohammad-2018-examining,may-etal-2019-measuring,gonen-goldberg-2019-lipstick-pig}; however they use synthetic sentence templates, while we extract naturally occurring sentences from the target corpus.

Our work is closely related to counter-factual token fairness
\cite{garg2019counterfactual}, which measures the magnitude of model prediction change when identity terms (such as \textit{gay}, \textit{lesbian}, \textit{transgender} etc.) referenced in naturally occurring sentences are perturbed. 
Additionally, \newcite{de2019bias} study gender bias in occupation classification using names in online biographies. 
In contrast, we propose a general framework to study biases with named entities.
\fromm{Noting that this can also be flipped to common names to introduce the same narrative, but with a focus on how people with the same names as public figures not be affected by the negativity associated to the individual.}
Our work is also related to the work on interpreting neural network models by manipulating input text \cite{li2016understanding}; while their aim is to interpret model weights, we study the model outputs for biases.

\fromm{Suggested replacement for below: However, if the proposed framework is altered so that it does not {\it perturb} named person entities within a slot and measure their differences, but instead measures the difference in the sentiment/toxicity expressed around the original named entities in-context, then
it is critical to model the linguistic associations from names. Otherwise, this task may be suitable for learning just that.}

\section{Discussion and Conclusion}
\label{sec_discussion}

Social biases towards real-world entities are often reflected in the language that mentions them, and such biases can be unintentionally perpetuated to trained models.  The focus of this paper is to introduce a simple method, Perturbation Sensitivity Analysis, to test for unwanted biases in an NLP model. 
Our method can be employed to study biases towards individuals (as demonstrated here), or
towards groups (e.g., races, genders), and is flexible enough to be applied across domains.

We are motivated to provide solutions for end users of NLP systems, who are likely to use models outside of their original training/testing environments, e.g., on data from populations or platforms that the system was not explicitly trained on.  The relative simplicity of the proposed approach suggests that the same method may be applied in different genres and across different languages, provided that a set of anchors are provided, such as pronouns in the target language.  Pronouns' status cross-linguistically as closed-class -- high frequency and easily listed as a small set of words -- make them particularly amenable for serving as a starting point for open domain bias analyses.

After identifying unwanted biases in a model, a next logical step is to reduce these biases.  Adapting the proposed approach to model training is straightforward, either by perturbing names in the training data directly, or by estimating the likelihood of given annotations as a function of sentence perturbation. Without access to model retraining, a simple solution could use post-processing to return system scores as a function of perturbed sentences, such as by averaging scores across perturbed sentences.

Future work could employ our method to study various group biases such as nationality, caste, and religion, since person names may function as significant markers for many such demographic associations. Our method could also be easily extended to other kinds of NLP models (beyond classification) as well as other types of entities (locations, organizations etc.).

\fromm{I suggest just removing the below at this point.  If there is room for it, then the content to add is in the tex comments. }

\paragraph{Acknowledgements} We would like to thank
the anonymous reviewers for their helpful and constructive feedback.
We also thank
Dylan Baker,
Emily Denton,
Yoni Halpern,
Ben Packer,
Lucy Vasserman,
Kellie Webster,
and
Simone Wu
for their valuable discussions on this paper.

\newpage

\bibliography{emnlp2019}
\bibliographystyle{acl_natbib}
\newpage

\end{document}